\documentclass[%
reprint,
superscriptaddress,
 amsmath,amssymb,
 aps,
 pre,
floatfix,
]{revtex4-1}

\usepackage{graphicx}% Include figure files
\usepackage{dcolumn}% Align table columns on decimal point
\usepackage{bm}% bold math
\usepackage{hyperref}% add hypertext capabilities
\usepackage{xcolor} 
\begin{document}

\preprint{APS/123-QED}

\title{Anomalous diffusion dynamics of learning in deep neural networks}% Force line breaks with \\

\author{Guozhang Chen}
\author{Cheng Kevin Qu}%
\author{Pulin Gong}%
\email{pulin.gong@sydney.edu.au}
\affiliation{%
 School of Physics, University of Sydney, Sydney, NSW 2006, Australia
}%

\date{\today}% It is always \today, today,

\begin{abstract}
Learning in deep neural networks (DNNs) is implemented through minimizing a highly non-convex loss function, typically by a stochastic gradient descent (SGD) method. This learning process can effectively find good wide minima without being trapped in poor local ones. We present a novel account of how such effective deep learning emerges through the interactions of the SGD and the geometrical structure of the loss landscape. We find that the SGD exhibits rich, complex dynamics when navigating through the loss landscape; initially, the SGD exhibits anomalous superdiffusion, which attenuates gradually and changes to subdiffusion at long times when approaching a solution. Such learning dynamics happen ubiquitously in different DNNs types such as ResNet and VGG-like networks and are insensitive to batch size and learning rate. The anomalous superdiffusion process during the initial learning phase indicates that the motion of SGD along the loss landscape possesses intermittent, big jumps; this non-equilibrium property enables the SGD to escape from sharp local minima. By adapting the methods developed for studying energy landscapes in complex physical systems, we find that such superdiffusive learning dynamics are due to the interactions of the SGD and the fractal-like regions of the loss landscape. We further develop a simple model to demonstrate the mechanistic role of the fractal-like loss landscape in enabling the SGD to effectively find global minima. Our results reveal the effectiveness of deep learning from a novel perspective and have implications for designing efficient deep neural networks.
% \begin{description}
% Secondary publications and information retrieval purposes.
% You may use the \texttt{description} environment to structure your abstract;
% use the optional argument of the \verb+\item+ command to give the category of each item. 
% \end{description}
\end{abstract}

%Use showkeys class option if keyword
\maketitle

%\tableofcontents

\section{Introduction}
Deep neural networks (DNNs) trained by stochastic gradient descent (SGD) have achieved great success in many application areas \cite{LeCun2015DeepLearning}.
As often assumed, the SGD optimizer of highly non-convex loss functions is rarely trapped in local minima, and effectively finds wide ones with good generalization \cite{Lipton2016StuckSpace,Hoffer2017TrainNetworks}.
Understanding how this property emerges from the DNNs is of fundamental importance for deciphering the secret of the remarkable effectiveness of deep learning \cite{Sejnowski2020TheIntelligence}.

\begin{figure*}
\centering
\includegraphics[width=\textwidth]{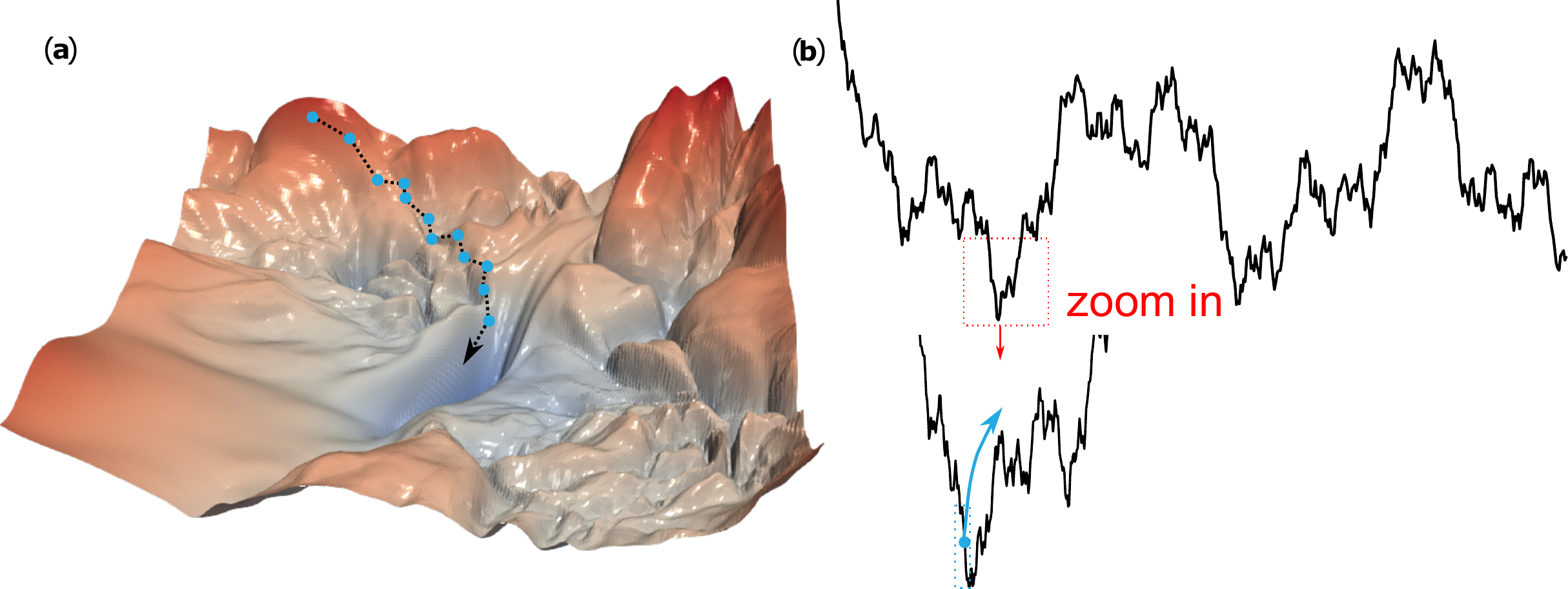}
\caption{\label{fig:sgd_schematic} \textbf{Schematic diagram of non-convex loss landscape with a fractal-like structure.}
(a) The log-value of loss landscape projected to 2D shows complex structures. The training process based on SGD can be regarded as a SGD optimizer moving on the loss landscape.
(b) The fine structure of non-convex loss landscape with a fractal structure shows self-affine and hierarchical properties.}
\end{figure*}

Recently, progress has been made in either characterizing the structure of loss functions or the dynamics of SGD for gaining comprehension of deep learning. For instance, loss functions have been studied by using random matrix theory \cite{Choromanska2015TheNetworks}, algebraic geometry methods \cite{Becker2018GeometryNetworks} and visualization-based methods \cite{Li2018VisualizingNets}. The dynamics of SGD have been examined via models of stochastic gradient Langevin dynamics with an assumption that gradient noise is Gaussian \cite{Jastrzebski2017ThreeSgd,welling2011bayesian}; in these models, the SGD is assumed to be driven by Brownian motion in particular. However, it has been increasingly realized that such Brownian motion-based characterizations of SGD dynamics are inappropriate, as SGD dynamics commonly exhibit highly anisotropic and dynamic-changing properties \cite{Chaudhari2018StochasticNetworks,Baity-Jesi2019ComparingSystems,Geiger2019JammingNetworks,Simsekli2019ANetworks}, suggesting the presence of rich, complex learning dynamics in DNNs. Recently, it has been shown that an inverse relation holds between the landscape flatness and the weight variance \cite{Fenge2015617118}. Despite the progress made by these studies, the fundamental questions of how the interaction of SGD with the structure of the loss function gives rise to complex learning dynamics, and whether and how such dynamics enable SGD to find wide minima remain unknown.\\
\indent In this study, by adapting the methods developed in nonequilibrium physical systems, we find that the interactions of the loss landscape and the SGD give rise to complex learning dynamics; these include anomalous superdiffusion during the initial learning phase, which changes to subdiffusion at long times when approaching a solution. During this process, the SGD walker moves from rougher (fractal-like) regions to flatter regions of the loss landscape. The fractal-like regions of the loss landscape indicate that they possess varying steepness (Fig.~\ref{fig:sgd_schematic}) and that the corresponding loss gradient displays large fluctuations with heavy-tailed distributions, thus resulting in superdiffusive learning dynamics. Superdiffusion consists of small movements that are intermittently interrupted by big jumps; these enable SGD to escape from local minima, thus effectively exploring the loss landscape. This computational advantage of superdiffusion is further illustrated in a simple model where the SGD interacts with a low-dimensional fractal loss landscape. Due to its movement even slower than a normal diffusive process (i.e. Brownian motion), the subdiffusive dynamics,  on the other hand, may consolidate the residence of the SGD in the flatter areas with good solutions.

\section{DNNs setup and characterizations of anomalous diffusion dynamics}
\subsection{DNNs setup}
We consider two classes of neural networks:
1) ResNet-14/20/56/110 \cite{He2016DeepRecognition}, where each type is labeled with the total number of layers it has. 
2) “VGG-like” networks that do not contain shortcut/skip connections. We produce these networks simply by removing the shortcut connections from ResNets, termed ResNet-14/20/56/110-noshort. 
All models are trained by vanilla SGD on multiple datasets including MNIST and CIFAR-10, by using two types of loss functions (i.e., cross-entropy, and Kullback Leibler divergence losses).
The training processes each run for 500 epochs.
All networks are initialized in the standard procedures of the PyTorch library (version 1.3.0).
Source code is available at \url{https://github.com/ifgovh/Anomalous-diffusion-dynamics-of-SGD.git}.

\subsection{Characterizations of anomalous diffusion learning dynamics}
Given that full-batch gradient descent is computationally expensive, in the SGD algorithm, the weight parameters $\mathbf{w} = (w_1,w_2,\cdots,w_d)$ are estimated by minimizing the minibatch loss function \smash{$\nabla \tilde{L}(\mathbf{w}) : \mathbb{R}^d \rightarrow \mathbb{R}$}, according to
\begin{equation}\label{eq:iteration_discrete}
\mathbf{w}_{t+1}=\mathbf{w}_{t} - \eta \nabla \tilde{L}_{t}\left(\mathbf{w}_{t}\right),
\end{equation}
where $\mathbf{w}_t$ denotes the $d$-dimension weight vector $(w_1,w_2,\cdots,w_d)$ at time $t$, and $\eta$ is the learning rate. 
The partial absence of the dataset generates gradient noise $U_t\triangleq\nabla \tilde{L}_{t}\left(\mathbf{w}_{t}\right) - \nabla L_{t}\left(\mathbf{w}_{t}\right)$.
The updating rule can be rewritten as:
\begin{equation}\label{eq:iteration_gradient_noise}
\mathbf{w}_{t+1}=\mathbf{w}_{t} - \eta \nabla L_{t}\left(\mathbf{w}_{t}\right) + U_t,
\end{equation}
Hence the SGD training process is also a random diffusive process, where $\mathbf{w}$ can be geometrically interpreted as coordinates of the SGD optimizer in the loss landscape $L$ (Fig.~\ref{fig:sgd_schematic}).

\begin{figure*}
\includegraphics[width=\textwidth]{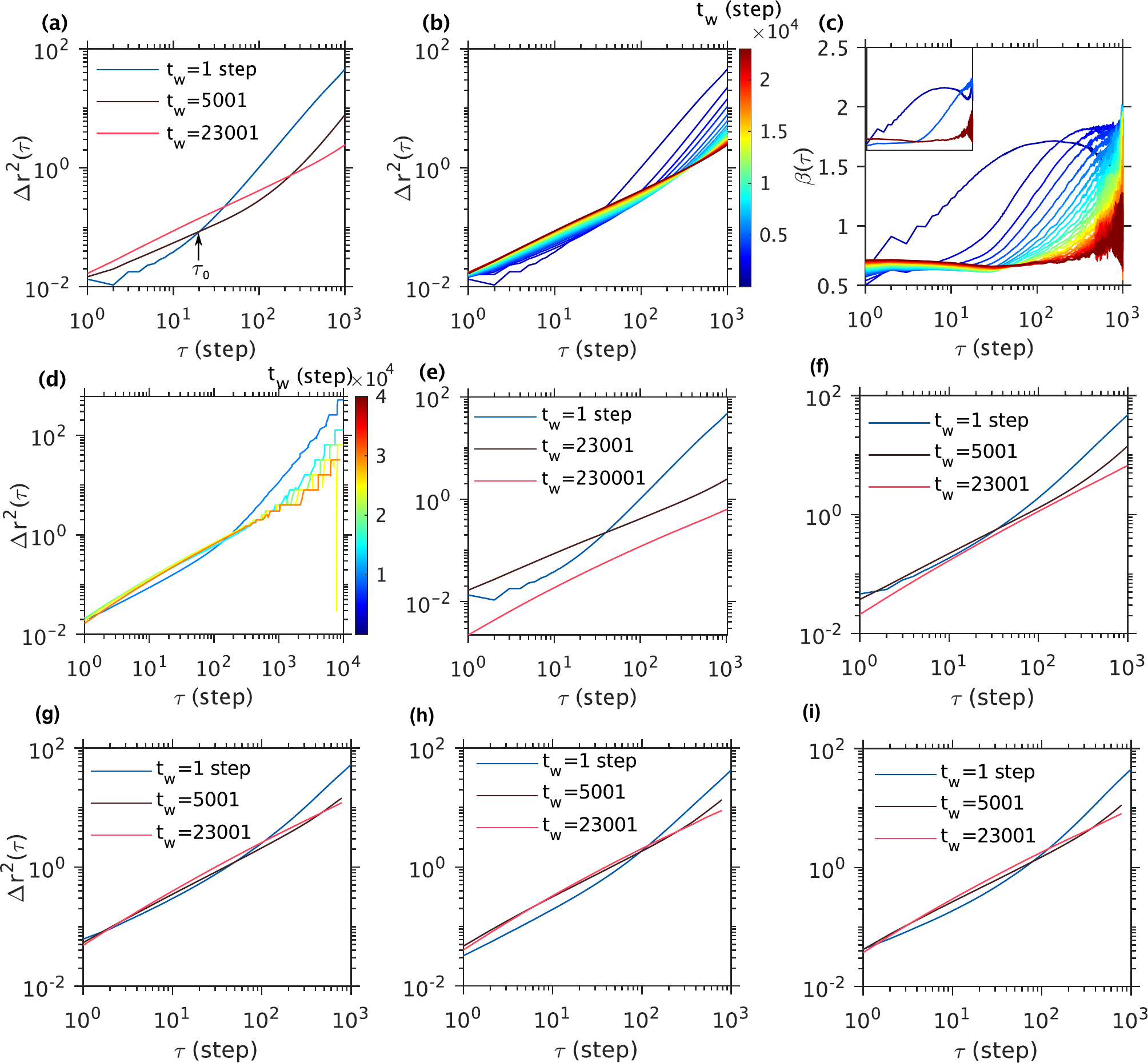}
\centering
\caption{\label{fig:superdiffusion} \textbf{SGD dynamics are anomalous diffusion.}
(a) MSD $\Delta r^2(\tau)$ of SGD as a function of lag time $\tau$ in interval $[t_w,t_w+T],\,T=1000$ for ResNet-14 (batch size of 1024, 500 epochs).
(b) Same as in (a) but the starting time of the interval $t_w$ gradually increases from 1 to 24000 steps, covering the whole training process.
(c) Logarithmic derivative $\beta(\tau)$ of the MSD shown in (b). Inset: Logarithmic derivative $\beta(\tau)$ of the MSD shown in (a). The color scheme is the same as in (b).
(d) Same as in (b) but for $T=10000$.
(e) Same as in (a) but for training ResNet-14 5000 epoch.
(f-i) Same as in (a) but for ResNet-110 (batch size: 512), ResNet-56 (batch size: 128), ResNet-20-noshort (batch size: 128), and ResNet-20 (batch size: 128), respectively.
}
\end{figure*}

The loss function of DNN exhibits complex structures, as demonstrated by the projected 2D loss landscape of ResNet-56-noshort \cite{Li2018VisualizingNets} (Fig.~\ref{fig:sgd_schematic}), analogous to complex energy landscape in physical systems \cite{Charbonneau2014FractalGlasses,Hwang2016UnderstandingApproach,Jin2017Exploringrheology,Cao2019PotentialRheology}.
In these physical systems, energy landscapes possess fractal-like structures and anomalous diffusion motions of particles stem directly from such kinds of structures.
In this study, we apply similar methods used in these systems to quantify the diffusion dynamics of the SGD optimizer.
Particularly, the time-averaged mean squared displacement (MSD) \cite{golding2006physical,bronstein2009transient} is used to characterize the dynamics of SGD moving through the loss landscape, which is defined as 
\begin{equation}\label{eq:MSD}
\Delta r^2(t_w,\tau) = \frac{1}{T}\sum_{t=t_w}^{t_w + T} \sum_{i=1}^{d}\left(w_{i}\left(t+\tau\right)-w_{i}\left(t\right)\right)^{2},
\end{equation}
where $\tau$ is lag time, $T$ is the length of the time interval $[t_w,t_w+T]$, and $w_{i}(t)$ is the value of the $i^{\text{th}}$ weight at time $t$.
$t_w$ is the time lapse after the start of the training process (i.e., waiting time). The time variables $t$, $t_w$, and $T$ are in units of iteration, and the unit time step corresponds to a single update of the weights.

We characterize the diffusion dynamics based on time-averaged MSD. Although ensemble-averaged MSD is the most appropriate in theory \cite{Zaburdaev2015LevyWalks}, averaging the ensemble of a high-dimension system is impossible in practice. Therefore, time-averaged MSD is a common tool used to quantify anomalous diffusive processes \cite{golding2006physical,bronstein2009transient}. 
No-averaged MSD (Eq.~\ref{eq:naMSD}) has been used to demonstrate how much the configuration of DNNs and spin glass models at time $t_w+\tau$ decorrelates from the one at time $t_w$ \cite{Baity-Jesi2019ComparingSystems}.
\begin{equation}\label{eq:naMSD}
\Delta(t_w,t_w + \tau) = \frac{1}{d} \sum_{i=1}^{d}\left(w_{i}\left(t+\tau\right)-w_{i}\left(t\right)\right)^{2}.
\end{equation}
However, we have calculated no-averaged MSD and found that no-averaged MSD curves are too noisy to characterize the anomalous diffusion learning dynamics.

Brownian motion is identified by $\Delta r^2(\tau) \propto \tau^{\alpha}$, for large $T$ with the diffusion exponent $\alpha=1$; the MSD is a linear function of lag time $\tau$.
When $\alpha\neq 1$, the corresponding diffusion process has a nonlinear relationship with respect to $\tau$ and is defined as anomalous diffusion \cite{Metzler2000TheApproach}.
If $1<\alpha<2$, it is a superdiffusive process; superdiffusion consists of small movements that are intermittently interrupted by long-range jumps.
This process has been widely observed in complex physical and biological systems \cite{Zaburdaev2015LevyWalks}; the mixture of small movements and big jumps in this process is essential for optimally transporting energy in turbulent fluid \cite{solomon1993observation}, and for animals to optimally search for spatially distributed food \cite{Viswanathan1999OptimizingSearches}.
If $0<\alpha<1$, the optimizer is subdiffusive, indicating that it moves slower on average than a normal diffusion process.

\section{Results}
\subsection{Anomalous diffusion of SGD dynamics}
We first illustrate that anomalous diffusion processes characterize SGD learning dynamics. As the findings of different DNN settings are similar, we thus demonstrate all results in ResNet-14 with a batch size of 1024, trained on CIFAR-10 with the cross-entropy loss function and a learning rate of 0.1, unless stated otherwise.

The MSD of each interval $[t_w,t_w+T]$ is calculated according to Eq.~\ref{eq:MSD} ($T=1000$) for a given $t_w$.
To demonstrate how the diffusion dynamics of SGD optimizer change during the training process, we change $t_w$ systematically. As shown in Fig.~\ref{fig:superdiffusion}(a), there are distinct regimes of the learning dynamics characterized by the diffusion exponent $\alpha$.
For the first regime $t_w < t_0$, $t_0=21000$ (blue curve, Fig.~\ref{fig:superdiffusion}(a)), MSD curves have two segments on the scale $\tau \in [1,T]$ with the smooth transitions around $\tau_0$ ($\tau_0$ is labeled in Fig.~\ref{fig:superdiffusion}(a)).
When the lag time $\tau>\tau_0$, the MSD curves can be fitted to $\Delta r^2(\tau) \propto \tau^{\alpha}$ with $\alpha>1$, indicating that the SGD optimizer exhibits superdiffusive dynamics.
However, when $\tau<\tau_0$, $\alpha<1$, i.e. the SGD dynamics are subdiffusive (Fig.~\ref{fig:superdiffusion}(a)).
The diffusion exponent $\alpha$ is calculated via the least-squared fitting method. We attempt to fit $\Delta r^2(t_w,\tau) \propto \tau^{\alpha}$ for $\tau \in [\tau',T]$. We determine $\tau' \in [0,T)$ as the smallest value such that the root-mean-square deviation RMSE $<0.03$; this $\tau'$ is denoted as $\tau_0$.

During the initial phase of the training process, the interval of the superdiffusion is much longer than that of the subdiffusion.
Nevertheless, as $t_w$ increases, the superdiffusion gradually attenuates, as demonstrated by the decrease of the diffusion exponent $\alpha$ and the increase of $\tau_0$ (brown curve in Fig.~\ref{fig:superdiffusion}(a); all curves are shown in Fig.~\ref{fig:superdiffusion}(b)).
In the second regime $t_w>t_0$, the diffusion exponent $\alpha=0.78$, i.e., the subdiffusion process eventually becomes dominant, as shown by the red curve in Fig.~\ref{fig:superdiffusion}(a). 
To identify the change from the first regime to the second one, we estimate $t_0$ by fitting the MSD curves ($[t_w,t_w+T]$). To do this, we shift $t_w$ from 1 to 23001, and $[t_0,t_0+T]$ is the first curve whose goodness of fit has RMSE $<0.03$ (0.03 is the standard deviation of all RMSE values).
These phenomena can be summarized as below:
\begin{equation}
\Delta r^2(t_w,\tau) \propto
\begin{cases}
\tau^{\alpha_1} \ \ \text{if} \ t_w < t_0, \tau < \tau_0 \\
\tau^{\alpha_2} \ \  \text{if} \ t_w < t_0, \tau \geq \tau_0\\
\tau^{\alpha_3} \ \  \text{if} \ t_w \geq t_0
\end{cases}    
\end{equation}
where $\alpha_1,\alpha_3 \in (0,1)$ and $\alpha_2 > 1$.
Such complex dynamics can be further quantitatively characterized by the logarithmic derivative of the MSD, $\beta$ \cite{dieterich2008anomalous,alves2016transient}, 
\begin{equation}
\beta(t_w,\tau)=\frac{\ln\,\Delta r^2(t_w,\tau+\Delta \tau) - \ln\,\Delta r^2(t_w,\tau)}{\ln\,(\tau+\Delta \tau) - \ln\,\tau},
\end{equation}
where ($\Delta \tau=20$, Fig.~\ref{fig:superdiffusion}(c)); in the first regime, $\beta$ quickly increases from a value smaller than 1 to a value larger than 1, but in the second regime, $\beta$ is larger than 1 only when $\tau>532$. 

The time-inhomogeneous anomalous diffusion dynamics are not sensitive to the interval $T$. For other values such as $T=5000$, $T=10000$, the SGD process exhibits similar dynamics with a change from a superdiffusion-dominated regime to a subdiffusion one. Figure~\ref{fig:superdiffusion}(d) shows an example of $T=10000$; initially, the SGD optimizer shows subdiffusion when $\tau<\tau_0$ (blue curve); gradually, the superdiffusion attenuates and subdiffusion process eventually emerges (red curve).
Note that we show the results in 500 epochs (24000 steps) because the dynamics after 500 epochs remain the same. The same model is also trained for 5000 epochs, the corresponding MSD curves for $t_w>24000$ still demonstrate subdiffusion (Fig.~\ref{fig:superdiffusion}(e)).

In addition, the time-inhomogeneous anomalous dynamics are not specifically unique to DNN models. Figures~\ref{fig:superdiffusion}(f-i) illustrate several models and they also demonstrate similar learning dynamics, including ResNet-14 with the batch size of 128, ResNet-14-noshort with the batch size of 1024, and ResNet-56 with the batch size of 1024.
These models are trained with a learning rate of 0.1.
This result thus indicates that the SGD learning dynamics generally possess an initial superdiffusion-dominated phase, which gradually evolves to a subdiffusion phase.

\begin{figure*}
\centering
\includegraphics[width=0.8\textwidth]{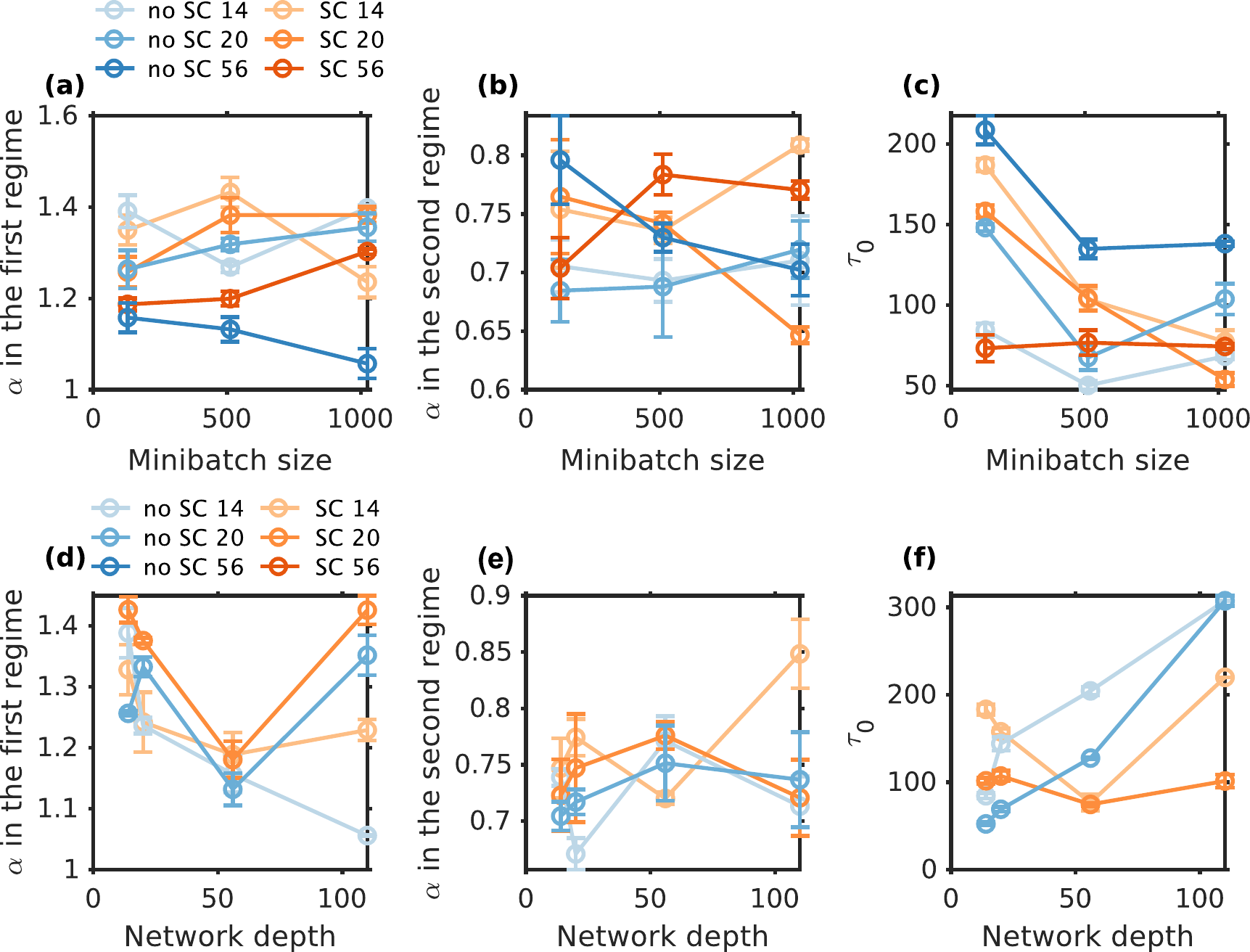}
\caption{\label{fig:width_batch} \textbf{The effects of the depth, batch size, and shortcut connections of DNNs on the anomalous diffusion dynamics of SGD.} The orange and blue colors represent the network with/without shortcut connections (SC) respectively.
The digits in legends of the first reow (a-c) represent network depth; for example, no SC 14 denotes ResNet-14-noshort. Those in the second row (d-f) represent minibatch size; for example, no SC 128 denotes ResNet training using the minibatch size of 128.
(a) The diffusion exponents $\alpha$ on larger lag times ($\tau>\tau_0$) when $t_w=1$ as a function of minibatch size.
(b) Similar to (a) but for $\alpha$ in the second regime (when $t_w=t_0$).
(c) The crossover $\tau_0$ as a function of minibatch size. Here, $\tau_0$ is the lag time when the MSD curve transitions from subdiffusion to superdiffusion when $t_w=1$. 
(d-f) Similar to (a-c) but for varying network depth.
The error bars represent the standard deviation calculated over 10 trials.}
\end{figure*}

\begin{figure*}
\centering
\includegraphics[width=0.9\textwidth]{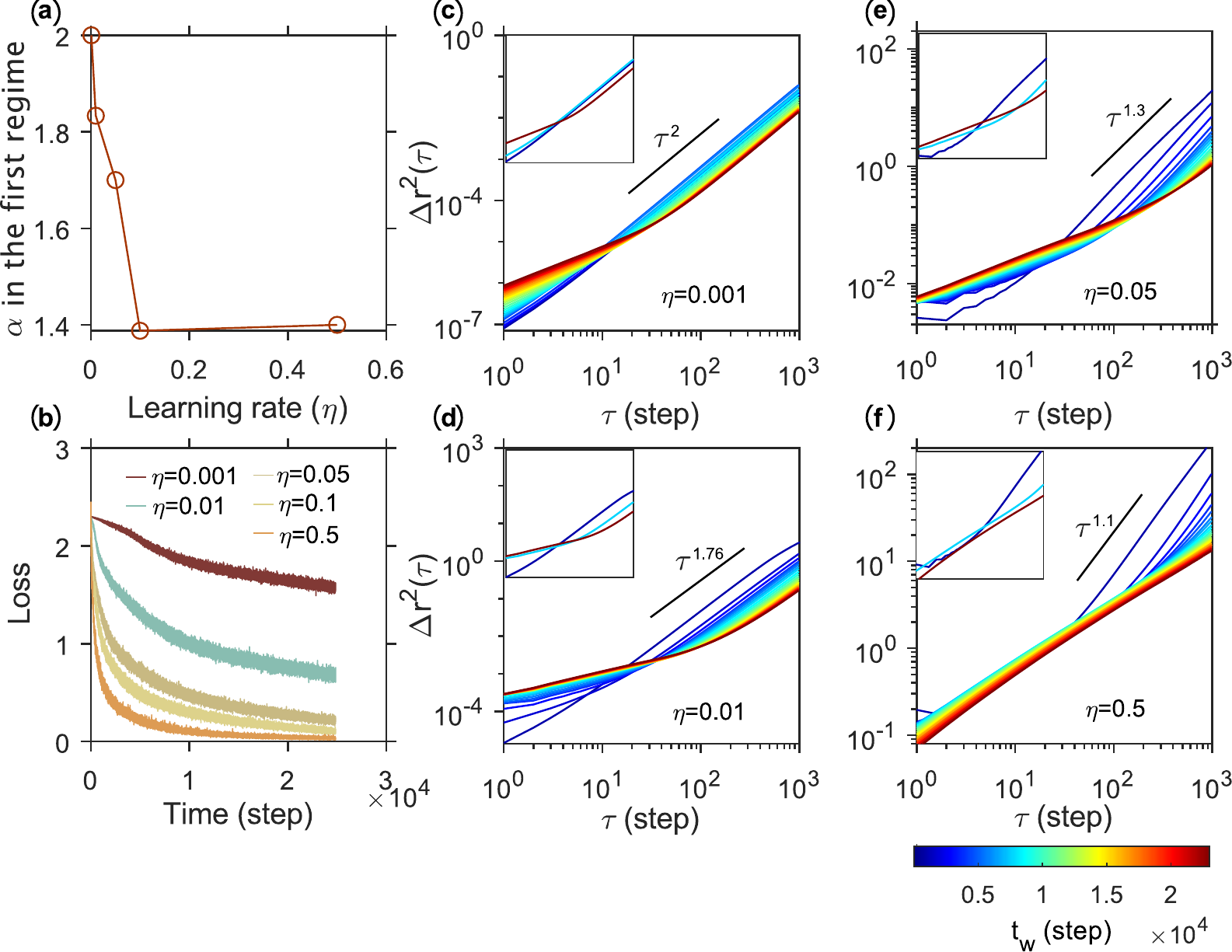}
\caption{\label{fig:lr} \textbf{The effect of learning rate on the anomalous diffusion dynamics of SGD.} All results are from ResNet-14 with a batch size of 1024 on CIFAR10 dataset. 
(a) The diffusion exponent $\alpha$ on larger lag times ($\tau>\tau_0$) when $t_w=1$ as a function of learning rate $\eta$.
(b) The loss value as a function of time.
(c-f) The MSD of SGD for the learning rates of 0.001, 0.01, 0.05, and 0.5, respectively. Jet colormap represents the starting time point ($t_w$) as in Fig.~\ref{fig:superdiffusion}(b). One curve represents MSD in 1000 steps. Inset: The MSD of SGD when $t_w=1, 7001, 23001$. The black lines are eye guides.}
\end{figure*}

The anomalous diffusion learning dynamics provide a way to characterize contributions of network structures and different hyperparameters to the training process.
To demonstrate this, we train two types of DNNs, i.e., ResNet and VGG-like networks.
VGG-like networks are produced simply by removing shortcut connections from ResNets.
As shown in Figs.~\ref{fig:width_batch}(a-b) and (d-b), shortcut connections do not change the diffusion exponent $\alpha$ significantly in both regimes, with $\alpha$ being greater than 1.
However, they do affect the scale range of superdiffusion characterized by $\tau_0$ when $t_w=1$.
$\tau_0$ indicates the crossover of MSD in the first 1000 steps.
Specifically, with shortcut connections, $\tau_0$ is smaller than those without shortcut connections, indicating that the scale of superdiffusion is elongated (Fig.~\ref{fig:width_batch}(c) and Fig.~\ref{fig:width_batch}(f)).
The superdiffusion dynamics enable the SGD walker to explore larger areas of loss landscape in a fixed time than normal diffusion and subdiffusion processes. From this perceptive, DNNs with shortcut connections can facilitate training, which is consistent with previous studies \cite{He2016DeepRecognition,Li2018VisualizingNets}.

The anomalous diffusion learning dynamics are not very sensitive to minibatch sizes.
As shown in Fig.~\ref{fig:width_batch}(a), the diffusion exponent $\alpha$ in the first regime does not significantly vary ($\pm 15\%$) with respect to the change of minibatch size from 128 to 1024 in all networks, although $\tau_0$ decreases in ResNet-14,20 with the increase of minibatch size (Fig.~\ref{fig:width_batch}(c)).

Furthermore, as shown in Fig.~\ref{fig:width_batch}(d), $\alpha$ changes nonlinearly with respect to the network depth.
However, the network depth reduces the scale range of superdiffusion, characterized by $\tau_0$ (Fig.~\ref{fig:width_batch}(f)); this result suggests that the deeper DNNs are more difficult to be trained \cite{Sankararaman2019TheDescent} due to the shorter scale of superdiffusion.

We next study the effect of the learning rate $\eta$ on the anomalously diffusive learning dynamics.
By varying learning rates from 0.001 to 0.5, we find that it only influences the emerging sequence of the superdiffusion learning dynamics (Fig.~\ref{fig:lr}(a)).
As shown in Fig.~\ref{fig:lr}(b), DNN training with small learning rates is much slower than that with large learning rates.
%thus, the training progress with small learning rates is delayed, compared with large learning rates.
Thus, a small or large learning rate can only slow down or speed up the training procedure, but does not change the fundamental occurrence of anomalous diffusion dynamics.
When the learning rate is small ($\eta<0.05$), during 500 epochs, the training processes remain in the first regime; there is no pure subdiffusion in Fig.~\ref{fig:lr}(c) and Fig.~\ref{fig:lr}(d).
For example, when $\eta=0.001$, the MSD curves for $t_w<5000$ have only one segment instead of two and a diffusion exponent of $\alpha=2$. This is because small learning rates delay the training process.
However, DNNs trained with larger learning rates grant superdiffusion in the first regime and pure subdiffusion dynamics in the second regime, as shown in Fig.~\ref{fig:lr}(e) and Fig.~\ref{fig:lr}(f).

% \begin{table}
%   \caption{Time-inhomogeneous MSD.}
%   \label{table:MSD}
%   \centering
%   \begin{tabular}{lll}
%     \toprule
% %    \multicolumn{2}{c}{Part}                   \\
% %    \cmidrule(r){1-2}
% & Training Regime: $t_w \leqslant t_0$    & Training Regime: ($t_w > t_0$) \\
%     \midrule
% $\tau<\tau_0$ & \qquad \qquad  $\alpha < 1$  & \qquad \qquad  $\alpha < 1$    \\
% $\tau>\tau_0$ & \qquad \qquad  $\alpha > 1$ & \qquad \qquad  $\alpha < 1$       \\
%     \bottomrule
%   \end{tabular}
% \end{table}
\begin{figure*}
\includegraphics[width=\textwidth]{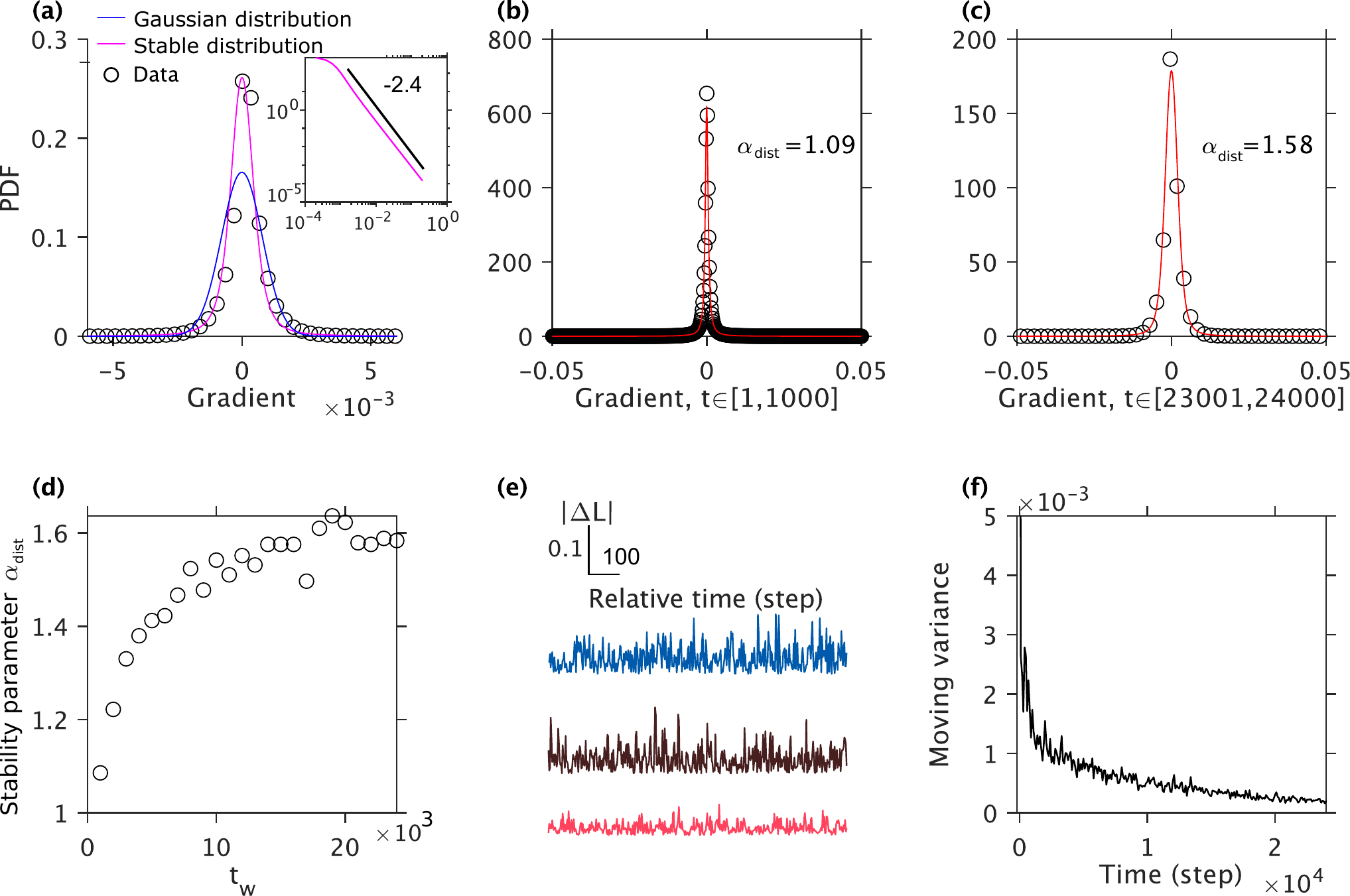}
\centering
\caption{\label{fig:heavy_tail_gradient} 
\textbf{The gradient is heavy-tailed.} The distribution of minibatch gradients $\nabla \tilde{L}(\mathbf{w})$ in the first regime can be fitted as a symmetric L\'evy $\alpha$-stable distribution (red curve). Inset: the log-log plot of the positive part of the distribution indicates that it has a power-law tail.
(b) When $t\in [1,1000]$, $\nabla \tilde{L}(\mathbf{w})$ can be fitted in symmetric L\'evy $\alpha$-stable distribution with the stability parameter $\alpha_{\text{dist}}=1.09$.
(c) Same as in (b) but for $t\in [23001,24000]$.
(d) The fitted stability parameter in the interval of $[t_w,t_w+T]$ as a function of $t_w$.
(e) Fluctuations of absolute value of change-of-loss $\Delta L$ decrease as $t_w$ increases. Coloarmap is the same as in Fig.~\ref{fig:superdiffusion}(a).
(f) The moving variance of loss as a function of time.
}
\end{figure*}

\subsection{Heavy-tailed gradients} 
To gain further insights into the physical origin of the anomalous diffusion dynamics, we next demonstrate the statistical property of minibatch gradients $\nabla \tilde{L}$. We first introduce the definition of the L\'evy $\alpha$-stable distribution.
%\textbf{As the SGD algorithm is directly realized by approximating the full batch gradient, it is important to understand the statistical properties of the minibatch. Even though gradient noise is the source of randomness, the minibatch gradient under certain conditions is an unbiased estimator of the full batch gradient [++], and hence it implicitly reflects upon on the structure of the loss landscape.??}
Given a L\'evy stable random variable $X$, it is characterized by the characteristic function \cite{Nolan2020}
\begin{equation}\label{eq:levy-char}
\varphi(u;\alpha_{\text{dist}},\beta,\gamma,\delta) =
%e^{iu\delta - \vert \gamma u \vert^{\alpha_{dist}}(1 - i \beta \text{sgn}(u)\Phi}
\exp\left( iu\delta - \vert \gamma u \vert^{\alpha_{\text{dist}}}(1 - i \beta \text{sgn}(u)\Phi) \right)
\end{equation}
where $\text{sgn}(u)$ is the sign of $u$ and
$$
\begin{aligned}
\Phi = 
\begin{cases}
\tan\left( \frac{\pi \alpha_{\text{dist}}}{2} \right) \quad &\alpha_{\text{dist}} \neq 1 \\
-\frac{2}{\pi}\log \vert t \vert &\alpha_{\text{dist}} = 1
\end{cases}
\end{aligned}
$$
$\alpha_{\text{dist}}$ is the stability parameter with the range $0<\alpha_{\text{dist}}<2$. The probability density function (PDF) decays with a power-law tail $|x|^{-\alpha_{\text{dist}} - 1}$ which is slower compared to Gaussian distributions; thus the distribution is heavy-tailed \cite{klafter2011first}. When $\alpha_{\text{dist}}=2$, the distribution is Gaussian.
$\beta$ is the skewness parameter. In particular, for a symmetric L\'evy $\alpha$-stable ($\mathcal{S} \alpha \mathcal{S}$) random variable $X$, i.e. $X \sim \mathcal{S} \alpha \mathcal{S}$, the skewness parameter $\beta = 0$ which indicates the PDF is symmetric around 0.
$\gamma$ is the scale parameter and $\delta$ is the shift parameter.
%Given a symmetric L\'evy stable random variable $X$, its probability density function (PDF) decays with a power-law tail $|x|^{-\alpha_{\text{dist}} - 1}$ which is much slower compared to Gaussian distributions, thus they have heavy tails \cite{klafter2011first}.

We next estimate the minibatch gradient \smash{$\nabla \tilde{L}$} in Eq.~\ref{eq:iteration_discrete} (vanilla SGD) with respect to each $w_i$ at each time point.
In the first regime ($t_w<t_0,\,t_0=21000$), it can be fitted to a symmetric\smash{ L\'evy} $\alpha$-stable distribution by the maximum likelihood method; stability parameter $\alpha_{\text{dist}}= 1.46528~[1.46509,1.46546]$ (the brackets denote the 95\% confidence interval).

The power-law tail of the distribution of \smash{$\nabla \tilde{L}$} shown in Fig.~\ref{fig:heavy_tail_gradient}(a) inset further validates the heavy-tailed distribution.
We further compare log-likelihood ratios between the fitted L\'evy $\alpha$-stable distribution and Gaussian distribution, and find that the log-likelihood ratios ($1.52\times10^9$) are sufficiently positive, indicating that the distributions most likely follow L\'evy $\alpha$-stable distribution ($p<10^{-15}$, Vuong test).

To illustrate the evolution of gradient distribution, we also fit the distributions of gradients in intervals $[t_w,t_w+T]$ to L\'evy $\alpha$-stable distribution; the log-likelihood ratios compared with Gaussian distribution are sufficiently positive. Figure~\ref{fig:heavy_tail_gradient}(b) demonstrates the distribution in the first interval, $t_w=1$; the stability parameter $\alpha_{\text{dist}}= 1.09~[1.05,1.12]$ . Figure~\ref{fig:heavy_tail_gradient}(c) demonstrates the distribution in the last interval, $t_w=23001$ and correspondingly $\alpha_{\text{dist}}= 1.58~[1.54,1.63]$. The stability parameter $\alpha_{\text{dist}}$ increases as training evolves (Fig.~\ref{fig:heavy_tail_gradient}(d)), indicating a reduction in the heavy-tailedness of gradient distribution. Because the heavier the tail, the larger the fluctuations of gradient values, and as the changes of gradient are directly related to the MSD (Eq.~\ref{eq:iteration_discrete} and Eq.~\ref{eq:iteration_gradient_noise}), the result regarding the changes of gradient distributions is consistent with the time-inhomogeneous anomalous diffusion dynamics where superdiffusion attenuates gradually to subdiffusion.
%, as the heavy-tailed gradient \smash{$\nabla \tilde{L}$}, according to Eq.~\ref{eq:iteration_discrete}, gives rise to long-range jumps ($\mathbf{w}_{t+1}-\mathbf{w}_{t}$) that result in superdiffusion processes.
It is also interesting to note that in the physics literature, it has been found that the increase of the heavy-tailedness of step sizes of random walkers results in super-diffusive motions \cite{Zaburdaev2015LevyWalks}. Such superdiffusive processes with intermittent long-range jumps might help the optimizer jump out local minima, facilitating basin hopping during the initial exploratory phase of training. This point is further illustrated below, based on a simple model of SGD.
%These long-range jumps caused by heavy-tailed gradients might help the optimizer jump out local minima.
Entering and leaving local minima give rise to the fluctuations of loss values ($L$).
To demonstrate the changes of these fluctuations, we first calculate the absolute value of change-of-loss $|\Delta L|=L(\mathbf{w}(t+1)-\mathbf{w}(t))$.
As shown in Fig.~\ref{fig:heavy_tail_gradient}(e), as $t_w$ increases, $|\Delta L|$ decreases.
Such behavior is further quantified by the decreasing moving variance of the loss $L$ against time (Fig.~\ref{fig:heavy_tail_gradient}(f));
the moving variance is calculated over a sliding window of 100 steps across neighboring $L$.

\subsection{Fractal trajectory of SGD}

\begin{figure*}
\centering
\includegraphics[width=\textwidth]{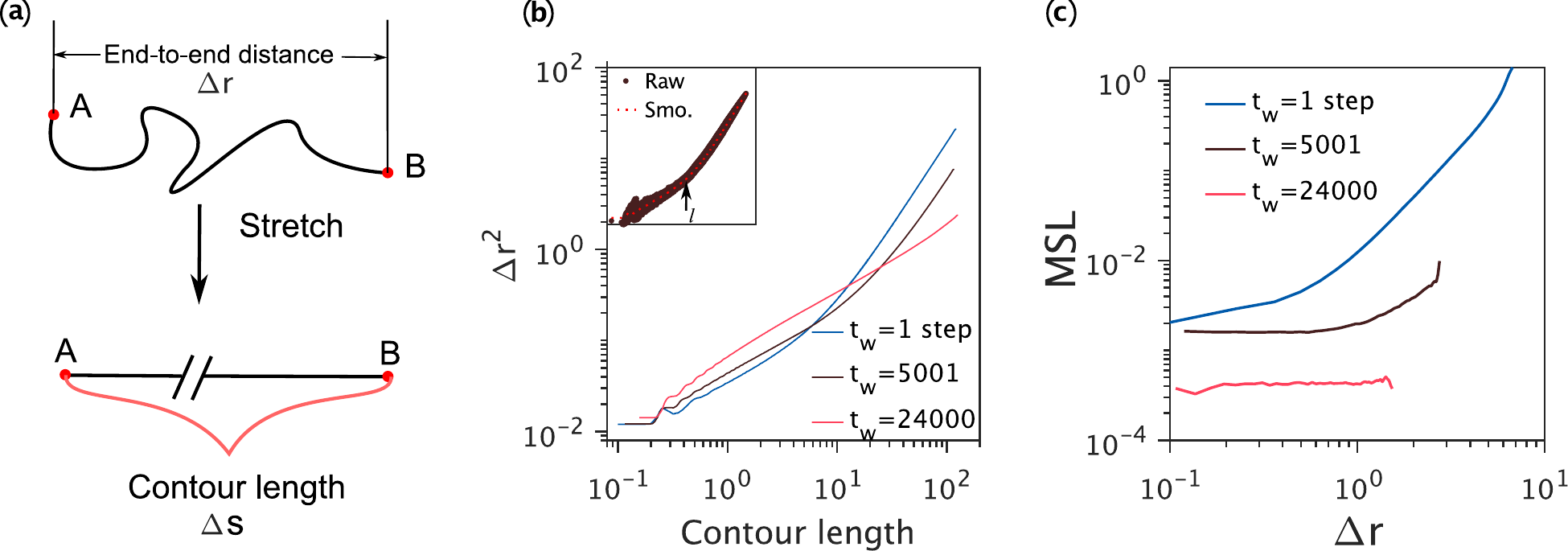}
\caption{\label{fig:fractal_landscape_path} \textbf{The fractal-like loss landscape of DNN and SGD path.} 
(a) Schematic diagram of contour length $\Delta s$ and end-to-end distance $\Delta r$.
(b) Squared end-to-end displacement ($\Delta r^2$) as a function of contour length.
The curve is smoothed by moving average over each window of raw data for clearer illustration; the window length is 40 steps. The inset displays the raw data when $t_w=1$ and the dashed red line represents the smoothed data.
(c) Mean squared loss differences of point pairs (MSL) against the end-to-end distances $\Delta r$ in the interval $[t_w,t_w+T]$.
The colormaps in (b) and (c) are the same. }
\end{figure*}

In complex physical systems, the anomalous diffusion of particles may associate with its fractal trajectories \cite{Hwang2016UnderstandingApproach}.
To verify that the SGD path is fractal, we characterize the power-law scaling of its end-to-end length and contour length as used in \cite{Hwang2016UnderstandingApproach}.
Specifically, the contour length is the path length from one end to another and is calculated by accumulating step sizes (curve length) as illustrated in Fig.~\ref{fig:fractal_landscape_path}(a).
As shown in Fig.~\ref{fig:fractal_landscape_path}(b), the segmented end-to-end distance ($\Delta r$) of SGD path appears to scale with its contour length, $\Delta s$. The data is smoothed to be clear;
the example of raw data and the smoothed data are shown in Fig.~\ref{fig:fractal_landscape_path}(b) inset.
The blue and brown curves in Fig.~\ref{fig:fractal_landscape_path}(b) display two distinct scaling regimes, illustrating the fractal property of SGD paths in certain ranges \cite{Hwang2016UnderstandingApproach}.
The fractal dimension $D_{\mathrm{f}}$ is calculated by the scaling of $\Delta s$ and $\Delta r$ ($\Delta r^{2} \sim \Delta s^{\lambda}$), $D_{\mathrm{f}}=(2 / \lambda) \in [1.32,2.67]$.
Separated by the crossover $l$ (labeled in Fig.~\ref{fig:fractal_landscape_path}(b) inset), $\lambda$ on longer length scales ($\Delta s>l$) is larger than that on short length scales ($\Delta s<l$).
As $t_w$ increases, MSL collapse to a power law with an exponent of 0.75 (red curve in Fig.~\ref{fig:fractal_landscape_path}(b)).
The time-inhomogeneous dynamical changes of SGD trajectory are similar to the case of the MSD (Fig.~\ref{fig:superdiffusion}(a)).
%and MSL (Fig.~\ref{fig:fractal_landscape_path}(a)).

To further determine whether the SGD path is self-affine or self-similar, we calculate the transverse distance and compare it with the end-to-end distance.
The transverse distance between two points on the trajectory is the maximal distance perpendicular to the straight line connecting these two points \cite{Hwang2016UnderstandingApproach}.
We find the transverse distances of different points along paths do not scale to their end-to-end distances; this indicates self-affine rather than self-similar fractal, because the former contains non-uniform scaling, i.e. the shapes are (statistically) invariant under transformations that scale different coordinates by different amounts \cite{Barnsley1988TheImages}.
%Set the trajectory and straight line that connects $w(t_1)$ and $w(t_2)$ as $\widehat{w_{t_1}w_{t_2}}$ and $\overline{w_{t_1}w_{t_2}}$ respectively, then the transverse distance for $w(t_1)$ and $w(t_2)$ is defined by the maximal distance perpendicular to $\overline{w_{t_1}w_{t_2}}$ from the curve $\widehat{w_{t_1}w_{t_2}}$.

\subsection{Fractal-like loss landscape}

\begin{figure}
\centering
\includegraphics{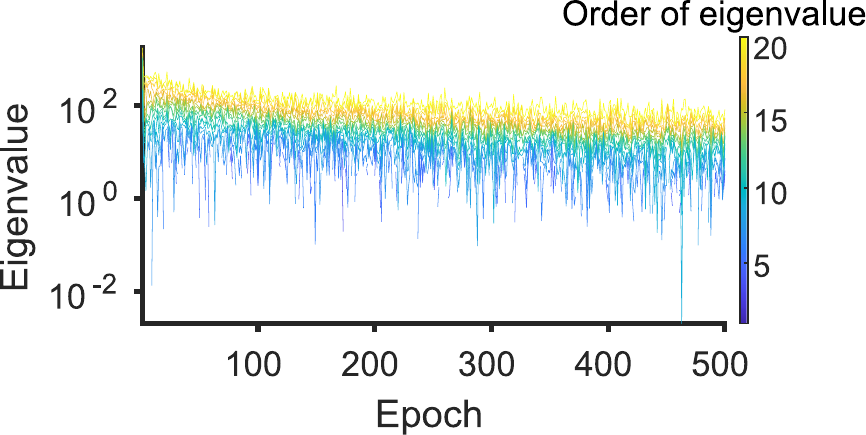}
\caption{\label{fig:hessian} \textbf{The top 20 eigenvalues of Hessian matrix of loss landscape decrease as training epoch increases.} This indicates that the SGD optimizer leaves narrow local minima and enters a flatter area in the loss landscape of DNN.}
\end{figure}

In physical systems, anomalous diffusion motions and fractal path of moving particles could be consequences of the energy landscape itself being fractal \cite{Charbonneau2014FractalGlasses,Hwang2016UnderstandingApproach,Cao2019PotentialRheology}.
Inspired by these studies, we hypothesize that the loss landscapes of DNNs could have fractal-like structures.
%\textcolor{red}{Inspired by these studies, we investigate the influence of different fractal setting on landscapes on the diffusion dynamics, and draw comparisons to that in the actual DNN.}
%\textcolor{red}{Different to Hessian matrix/volume based arguments, we look at a completely different quantity which characterizes roughness of landscapes: fractal dimension.}
It is impossible to directly quantify the fractal dimensions of high-dimensional loss landscape; we thus use an approach proposed in \cite{Sorkin1991EfficientLandscapes}.
From \cite{Barnsley1988TheImages}, the definition of fractal is as the following:\\
\textbf{Definition}
A random function $f$ on a metric space is fractal if the distribution of $f(X')$ conditional on $f(X)$, $X$, and $X'$, is normal with mean 0 and variance proportional to $\Delta r(X', X)^{2H}$, where $H$ is a parameter in $(0, 1)$ and $\Delta r(X', X)$ is the distance between point $X$ and $X'$.

The distribution of $X$ and $X'$ in the definition is over the probability space from which $f$ is drawn, but experimentally we sample over random values of $X$ and $X'$ for a given sample function $f$.
The equivalence of these two procedures is referred to as ergodicity and we take it for granted.
Checking the distributions for each value of distance in practice is difficult; hence we measure the expectation of $[f(X') - f(X)]^2$ to characterize the fractal-like structure and check if it satisfies Eq.~\ref{eq:fractal_criteria}.
In the context of a SGD optimizer on the loss landscape, $f$ is the loss function $L$; $X$ and $X'$ are a pair of weights ($\mathbf{w}$ and $\mathbf{\widetilde w}$) on the loss landscape.
Thus, the expectation is referred to as mean squared loss (MSL),
\begin{equation}\label{eq:fractal_criteria}
\text{MSL} \propto \Delta r\left(\mathbf{w}, \mathbf{\widetilde w}\right)^{2 H}
\end{equation}
where \smash{$\Delta r\left(\mathbf{w}, \mathbf{\widetilde w}\right)^{2 H}$} is the end-to-end distance between $\mathbf{w}$ and \smash{$\mathbf{\widetilde w}$}.
Note that in the field of machine learning, the same method has been used to quantify fractal landscapes for simulated annealing \cite{Sorkin1991EfficientLandscapes}.

The MSL is calculated by the following equation,
\begin{equation}
    \text{MSL}(t_w,\Delta r)=\frac{1}{\mathcal{N}_{\Delta r}}\sum_{j=1}^{\mathcal{N}_{\Delta r}} [L(\mathbf{w}^{t_w,T}_{\Delta r}(j))-L(\mathbf{\widetilde w}^{ t_w,T}_{\Delta r}(j))]^2,
\end{equation}
where the pair of weights $\mathbf{w}^{t_w,T}_{\Delta r}(j)$ and $\mathbf{\widetilde w}^{t_w,T}_{\Delta r}(j)$ are sampled at different time steps along the trajectory in $[t_w,t_w+T]$ with the end-to-end distance between them equal to $\Delta r$, and $\mathcal{N}_{\Delta r}$ is the total number of pairs at a distance of $\Delta r$. To be consistent with piecewise MSD, we choose $T=1000$ steps.
We use the points sampled by SGD (i.e., points along the optimization trajectory) to estimate MSL; it illuminates certain structure on the local area of the loss landscape that we would like to explore. 

As shown in Fig.~\ref{fig:fractal_landscape_path}(c) (blue curve), in the first regime ($t_w<t_0$), the MSL curve can be fitted to a power-law function with an exponent of 1.8 on the larger distance scale ($\Delta r \in[0.4,10]$). It satisfies Eq.~\ref{eq:fractal_criteria} and indicates that the loss landscape of DNN has fractal-like structures at the initial phase of learning process.
%analogous to efficient simulated annealing on fractal loss landscapes \cite{Sorkin1991EfficientLandscapes}.
Note that the power-law scalings do not hold within the whole scale ([0.1,10]).
As the superdiffusion attenuates, the end-to-end distance $\Delta r$ in $T$ decreases and the power-law exponent of the MSL with respect to $\Delta r$ flattens from period to period (brown curve in Fig.~\ref{fig:fractal_landscape_path}(c)). Eventually, in the second regime ($t_w>t_0$), the MSL is around a constant value against varying $\Delta r$ (red curve in Fig.~\ref{fig:fractal_landscape_path}(c)). Based on the above definition, this indicates that the optimizer now reaches a relatively flat region on the landscape. We use “flat” colloquially to indicate approximate flatness \cite{Hochreiter1997FlatMinima,Sagun2018EmpiricalNetworks,Baldassi161,Fenge2015617118}.
Our results thus indicate that the SGD optimizer moves from rougher (more fractal-like) to relatively flatter regions of the loss landscape.  The change to the flatter regions can also be quantitatively demonstrated by the fact that eigenvalues of the Hessian matrix gradually decrease to near-zero values (Fig.~\ref{fig:hessian}), which has also been found in \cite{pmlr-v97-ghorbani19b}. Note that there has been increasing evidence showing that the optimizer can eventually find a good generalizable solution existing at the flat regions of the loss landscape \cite{Hochreiter1997FlatMinima, Baldassi161, Chaudhari_2019}.
Our work suggests that it would be relevant to use the methods in these previous studies to characterize how the SGD moves from rougher to flatter regions of the loss landscape.
%There are also other measures of flatness quantified by the volume of basin of attraction. In particular, the volume from good minima dominates that of poor minima. \cite{2017arXiv170610239W}
%Importantly, these time-inhomogeneous behaviors are consistent with the anomalous diffusion of SGD.
%When $t_w<t_0$, the fractal-like hierarchical structure may provide heavy-tail gradients and thus superdiffusion emerges; when $t_w>t_0$, the MSL converges to a constant, potentially because the optimizer enters a flat basin eventually. \textcolor{red}{The fractal-like structure of the landscape is only present at the start of training, but not towards the end of training.}

%In total, our results deomnstrate that the SGD optimizer moves from rougher (fractal-like) to relatively flatter regions of the loss landscape.
%Importantly, these changes within loss landscape are consistent with those of the anomalous diffusion of SGD. 
Importantly, we find that when $t_w < t_0$, the fractal-like hierarchical structure provides highly fluctuating gradients and thus superdiffusion emerges; when $t_w > t_0$, the flatter structure causes fewer fluctuations of gradients and results in subdiffusion. In a recent study \cite{Fenge2015617118}, it has been shown that during the training process of DNNs, the SGD moves towards flatter regions of the loss landscape and correspondingly the anisotropic SGD noise strength decreases. In future work, it would be interesting to explore whether the changes from superdiffusion to subdiffusion underlie  the changes of noise strength as reported in \cite{Fenge2015617118}.

To further justify the fractal-like structure of loss landscape, we use the method of filter-wise normalized directions \cite{Li2018VisualizingNets} to project the loss landscape of ResNet-14 (batch size of 1024, learning rate of 0.1, trained on CIFAR-10 with cross-entropy loss function) to 2D space.
We then calculate the Hausdorff dimension of a 2D projected loss landscape via the box-counting method \cite{li2009improved}; the fractal dimension is approximately 1.8.

\subsection{Fractal-like landscapes can cause anomalous diffusion learning dynamics}
\begin{figure*}
\centering
\includegraphics[width=\textwidth]{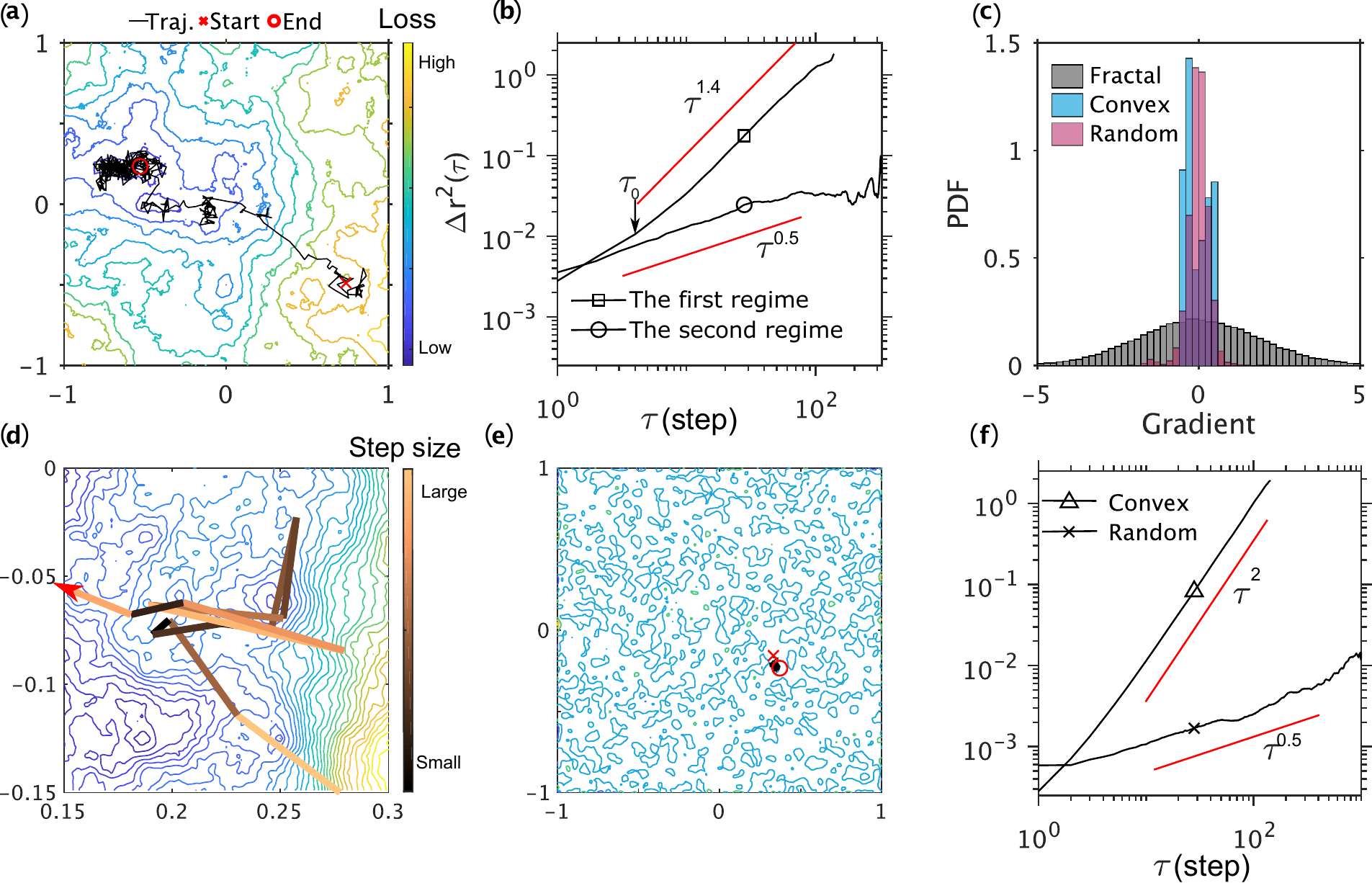}
\caption{\label{fig:toy_model} \textbf{The simplified model unravels the anomalous diffusion nature of the SGD optimizer.}
(a) Black curve represents the trajectory of the SGD optimizer in the simplified model on the 2D fractal loss landscape illustrated by the contour plot.
(b) MSD of the SGD optimizers in the fractal landscape as the function of lag time. The curve is averaged over 10 trials.
(c) Distributions of gradients ($\Delta L$) in fractal, convex, and randomly shuffled landscapes, respectively.
(d) Trajectory of the SGD optimizer and the landscape in (a) are zoomed in around a local minimum. The colormap from black to golden encodes the step sizes.
(e) Same as in (a) but for random landscape.
(f) Same as in (b) but for random landscape and convex landscape.
}
\end{figure*}

To further understand the contributions of fractal-like loss landscapes to the anomalous diffusion of learning dynamics, we develop a simplified model of SGD with a 2D fractal loss landscape, as described below.
%\textcolor{red}{We develop the following model involving a 2D fractal landscape to simulate the learning/optimization dynamics purely from the perspective of anomalous diffusion.}
\begin{equation}\label{eq:discrete_sgd_normal}
  \mathbf{w}_{t+1}=\mathbf{w}_{t}-\eta \nabla L\left(\mathbf{w}_{k}\right)+ \eta \sigma Z_t,
\end{equation}
where $\mathbf{w}_t$ is the weight parameters at time $t$, $\eta$ is the learning rate, and $Z_t$ is drawn from Gaussian distribution $\mathcal{N}\left(\mathbf{0}, \sigma \right)$.
The landscapes of $L$ are fractal on a 2D space and are generated by the algorithm in \cite{Douillet2020FractalGenerator} with $\text{fractal dimension} \in [1,2]$ (Fig.~\ref{fig:toy_model}(a)).
Among the generated landscapes, those which contain a wide minimum are selected, simulating the flat basin (local optima) in loss landscapes of DNNs. In such landscapes, the SGD walker moves from rougher regions to flatter regions of the loss landscape, as observed in DNNs. we iterate Eq.~\ref{eq:discrete_sgd_normal} 1000 times with the learning rate $\eta\in [0.002,0.01]$ and $\sigma\in [0.005,0.05]$, and the gradient $\nabla L$ is calculated by the numerical gradient of the landscape.
The SGD optimizer moves from a high-altitude point to the global minimum, as the example ($\eta=0.02$ and $\sigma=0.01$) shown in Fig.~\ref{fig:toy_model}(a). The MSD also illustrates that the SGD optimizer is dominantly superdiffusive when $t_w<t_0=139$ (before entering the final minimum; curve with a square in Fig.~\ref{fig:toy_model}(b)).
When $t_w>t_0$ (after entering the final minimum), the dynamics of SGD optimizer are only subdiffusive as shown in the curve with a circle in Fig.~\ref{fig:toy_model}(b).
Such behaviors demonstrate that the simple model can reproduce the time-inhomogeneous MSD of SGD as found in DNNs.
% To demonstrate a more realistic case of our 2D simple model, we also use the 2D projected loss landscapes of ResNet-14,56 \cite{Li2018VisualizingNets}.
% As the fractal-like structure of projected loss landscape, the anomalous diffusion process also manifests as observed in the full deep-learning scenarios.

In this model with 2D fractal landscape, the fractal landscape generates a heavy-tailed distribution of gradients ($\nabla L$) of all steps (Fig.~\ref{fig:toy_model}(c)).
As found in the DNNs, the gradient distribution can be fitted as a symmetric L\'evy $\alpha$-stable distribution which has the stability parameter $\alpha_{\text{dist}} =1.9~[1.88749, 1.91298]$.
The goodness of fit is verified by the positive log-likelihood ratio (39.37) of L\'evy $\alpha$-stable distribution and normal distribution.
In our model, $\nabla L$ is calculated by the numerical gradient of the fractal landscape when $t_w<t_0$.
Such heavy-tailed gradients provide a relatively higher possibility of long-range jumps, which generate superdiffusion.
In the fine structure of fractal landscape, there are large gradient values (Fig.~\ref{fig:sgd_schematic}(b)) which propel the SGD optimizer to jump out narrow minima.
It is important to note that as noise in our simplified SGD model is Gaussian, such heavy-tailed gradients and superdiffusion dynamics solely result from the fractal-like loss landscape.

To explicitly illustrate the benefits of fractal landscapes for facilitating the SGD to jump out local minima, we focus on some regions around local minima.
As shown in Fig.~\ref{fig:toy_model}(d), the optimizer moves to the local minimum where the SGD optimizer displays short-range movements as illustrated by darker bars in Fig.~\ref{fig:toy_model}(d) and long-range movements as illustrated by lighter bars in Fig.~\ref{fig:toy_model}(d).
Some long-range steps make the SGD optimizer escape the minimum; however, some of them jump to a lower altitude and then leave the minimum.
This example illustrates how the fractal landscape assists the SGD optimizer escape local minima.
As we only use Gaussian noise in this simple model, the main source providing long-range jumps to escape local minima is the heavy-tailed gradients (Fig.~\ref{fig:toy_model}(c)). This result suggests that in DNNS, the similar mechanism might enable the SGD walker to escape local minima in the initial superdiffusion-dominated regime. %supporting the observations in DNNs. 
 
Although the results in the simple model are largely consistent with DNNs, there are some differences. In the simple model, if the learning rate is small ($< 0.002$), the toy model becomes sensitive to initial conditions (the initial position of random optimizers). The random optimizer would be trapped in a local minimum. For large learning rates ($> 0.01$), combining with the occasional long-range jump, the optimizer would easily go out the landscape.

% We now explain why the MSD in the first regime has two segments. Compared with the large displacement outside local minima (step A and D in Fig.~\ref{fig:toy_model}(d)), the SGD optimizer is stuck transiently and the diffusive distance is small within $\tau<\tau_0$ ($d(\tau<\tau_0)$ in Fig.~\ref{fig:toy_model}(d)). When we calculate the MSD within $\tau<\tau_0$, the minority of large displacement outside the local minima is canceled by the majority of short displacement inside the local minima, making the MSD $\text{exponent} < 1$. When $\tau>\tau_0$, the optimizer jumps outside the minima ($d(\tau>\tau_0)$ in Fig.~\ref{fig:toy_model}(d)) and the long-range jumps turn into majority, making the MSD $\text{exponent} > 1$.
However, if we choose other types of landscapes and maintain $\eta$ and $\sigma$, the complex MSD dynamics no longer hold. 
When the landscape is generated by smoothing a randomly shuffled fractal landscape with a Gaussian kernel (standard deviation: 8) such that it is at least once-differentiable, the optimizer gets stuck in a local minimum (Fig.~\ref{fig:toy_model}(e)) and exhibits subdiffusion (Fig.~\ref{fig:toy_model}(f)).
On the other hand, when the landscape is convex, for example, a convex paraboloid, the MSD has an exponent close to 2 (Fig.~\ref{fig:toy_model}(f)), inconsistent with the results in DNNs. In comparison to the tail of gradient distribution in fractal landscapes, the ranges of gradients in convex or random landscapes are far smaller (Fig.~\ref{fig:toy_model}(c)). These results thus indicate the fractal-like loss landscape is essential for the emergence of the anomalous diffusion learning dynamics.
% Although the results in the simple model are largely consistent with DNNs, there are some differences. In the toy model, if the learning rate is much smaller, the toy model becomes sensitive to initial conditions (the initial position of random optimizers). The random optimizer would be trapped in a local minimum. The DNN, however, is not heavily affected by the initial conditions, potentially due to the reason that in the high dimensional landscape, the initial conditions are uniformly equivalent.

\section{Discussion}

We have revealed the anomalous diffusion nature of deep learning dynamics which arises from the interactions of the SGD walker with the geometry structure of the loss landscape. Particularly, we have demonstrated that the fractal-like loss landscape can give rise to the superdiffusion learning dynamics with intermittent big jumps during the initial training phase, which plays an essential role in preventing the SGD optimizer from being trapped in narrow minima. Subdiffusion on the other hand also occurs naturally during the final stage of the training process, stabilizing the movement of the optimizer gradually, potentially when wide minima of landscape are encountered. In addition, we have developed a new SGD model to reveal the mechanistic relations between the fractal landscape, the superdiffusive learning dynamics and their computational benefits.
Our results reveal the effectiveness of deep learning from the perspective of its rich, complex dynamics and have implications for designing efficient deep neural networks.

Previous studies tackled the problem from modeling the SGD as a process with heavy-tailed behaviors \cite{Martin2019TraditionalModels,Simsekli2019ANetworks,Zhang2019WhyModels}.
Particularly, Simsekli et al reported a heavy-tailed behavior in the stochastic gradient noise ($U_t$ in Eq.~\ref{eq:iteration_gradient_noise}) and proposed modeling the SGD dynamics as a stochastic differential equation driven by an alpha-stable process. They further invoked existing metastability theory to justify why these dynamics would prefer wide minima \cite{Simsekli2019ANetworks}. The SGD updating rule can be represented in terms of gradient noise, i.e., $\mathbf{w}_{k+1}=\mathbf{w}_{k}-\eta \nabla L\left(\mathbf{w}_{k}\right)-\eta\cdot$gradient noise.
However, the distribution of gradient noise was further found to be Gaussian in the early phases, and changes throughout the training process \cite{Panigrahi2019Non-gaussianityNoise}. To model the gradient noise as a single type of noise is thus inappropriate.
Rather than the gradient noise, our work focuses on the change of the drift term throughout the training process (i.e., gradient, $\nabla L\left(\mathbf{w}_{k}\right)$) which is directly related to the structure of loss landscape.

The structure of loss landscape can give rise to the anomalous diffusion dynamics of learning process quantified from MSD \cite{Hwang2016UnderstandingApproach}.
In \cite{Baity-Jesi2019ComparingSystems}, no-averaged MSD was used to analyze learning dynamics and the slope of MSD is not equal to 1. However,  this study did not introduce anomalous diffusion in the discussion of learning dynamics but only showed the aging phenomena. Ideally, we would like to obtain the ensemble average of the MSD, but this is unrealistic for DNNs.
We instead used the time-averaged MSD \cite{golding2006physical,bronstein2009transient,metzler2014anomalous,Zaburdaev2015LevyWalks,alves2016transient,grebenkov2019time}, which can better demonstrate the anomalous diffusion in the context of DNNs.

By using a similar methodology as in simulated annealing \cite{Sorkin1991EfficientLandscapes}, we have quantitatively demonstrated that the loss-landscape of DNN is fractal-like.
The fractal-like structure can give rise to heavy-tailed gradients which may help the SGD optimizer to jump out local minima.
Also, the fractal landscape may result in fractal trajectories of the SGD optimizer, as in complex physical systems \cite{Hwang2016UnderstandingApproach}.
Our results show that the SGD trajectories are indeed fractal as quantified by the contour length and end-to-end length.
Recently, it has been suggested that the fractal trajectory of SGD optimizer may facilitate generalization in DNNs \cite{imekli2020hausdorff}.
Such fractal trajectories might result from the fractal-like structure of loss landscape as what we have demonstrated. Our results based on both the MSD and MSL indicate that the SGD walker moves from rougher (more fractal-like) areas to flatter areas of the loss landscape. During this process, the learning dynamics change from superdiffusion-dominated dynamics, which help the SGD to escape from local traps, to subdiffusive dynamics which can rather consolidate the residence of the SGD in the flatter areas with good solutions (minima). These time-inhomogeneous anomalous diffusion learning dynamics arising from the interactions of SGD and the loss landscape thus provide insights into understanding how the optimizer can find flat minima.

The simple SGD model on a 2D fractal landscape generates the same pattern of time-inhomogeneous learning dynamics as in the high-dimensional DNNs, 
which however cannot be accounted for by the traditional formulation based on the Langevin equation with Gaussian noise \cite{Jastrzebski2017ThreeSgd, welling2011bayesian}.
%suggesting shared properties between the 2D simple model and the high-dimensional DNN. \textcolor{red}{Our model is not directly related to a DNN setting, but only shares certain similarities with respect to anomalous diffusion of the learning dynamics. We attempt to contrast the diffusion dynamics on landscapes settings with different degress of roughness, quantified by the fractal dimension.}
The simple SGD model does not involve any type of non-Gaussian noise and demonstrates that fractal landscapes alone can lead to anomalous diffusion learning dynamics, thus indicating that the interactions between the fractal loss landscape and SGD are the mechanism underlying the emergence of the anomalous diffusive learning dynamics. 
However, as our 2D model is not directly derived from DNN models, the generalization of this mechanism to DNNs is limited.
%the emergence of heavy-tailed noise may further facilitate anomalous diffusion dynamics. 
Anomalous superdiffusion and subdiffusion are nonlinear diffusive processes and are generally referred to as fractional motions that can be formulated based on fractional differential equations \cite{Metzler2000TheApproach}, suggesting that developing a fractional mean field theory as in \cite{wardak2021fractional} for understanding deep neural networks would be a promising direction to pursue in the future.
In addition, future studies should figure out the major source of fractal-like loss landscape. 
The training landscape is composed of the data and the network architecture.
Some previous studies have shown that realistic datasets such as handwritten digits (MNIST), rather than random noise, have low-dimension structure/manifold \cite{Costa2004LearningDatasets,Levina2004MaximumDimension,Goldt2019ModellingModel}.
It would be interesting to study the effect of such data structure on the geometrical properties of loss landscape.

On the other hand, the network structure can affect anomalous diffusion learning dynamics.
We have found that the deeper DNNs, the shorter scale of superdiffusion, indicating a more demanding training process, consistent with the empirical rules of DNNs \cite{Sagun2018EmpiricalNetworks,Santurkar2018HowOptimization,Sankararaman2019TheDescent} and extended the understanding from the aspects of training dynamics and landscape structures \cite{Geiger2019JammingNetworks}.
Additionally, shortcut connections in ResNet can extend the scale of superdiffusion, explaining why employing such techniques reduces the difficulties of training DNNs.
These findings agree with the theoretical and experimental results of gradient confusion \cite{Sankararaman2019TheDescent} and the visualization of 2D projected loss landscapes \cite{Li2018VisualizingNets}.
These studies found that shortcut connections reduce the difficulty in the training process by smoothing out the loss landscape.
Furthermore, we find that the batch size cannot significantly alter the anomalous diffusion dynamics, supporting the conclusion that gradient noise is not the only driving force to escape critical points (saddle points or local minima). 
For future studies, it remains important to find out the quantitative relation between other architectures such as batch normalization \cite{Santurkar2018HowOptimization} and fractal-like landscape structure.
\vspace{\baselineskip}
\vspace{\baselineskip}

\begin{acknowledgments}
%\section{Acknowledgements}
The authors acknowledge the University of Sydney HPC service for providing high-performance computing that has contributed to the research results reported within this paper. This work was supported by the Australian Research Council (grant nos. DP160104316, DP160104368).
\end{acknowledgments}

\bibliography{references.bib}% Produces the bibliography via BibTeX.

\end{document}